# Applying Deep Learning for cockpit segmentation in the context of mixed reality

**Alexandre Leles Sousa** * **Pedro de Oliveira Nielson** *
**Erick Oliveira Rodrigues** ** **Rafael Francisco dos Santos** *
**Giovani Bernardes Vitor** *

* *Laboratório de Robótica, sistemas inteligentes e Complexos - RobSIC*
*Instituto de Ciências Tecnológicas*
*Universidade Federal de Itajubá, Campus Itabira, MG.*
*(e-mail:giovanibernardes@unifei.edu.br)*
** *Universidade Tecnológica Federal do Paraná - UTFPR, Campus Pato Branco/PR*

**Abstract:** Computer vision is an area that has been growing continuously. With the advance of technologies with a first-person view, new development opportunities have emerged inside the area. Mixed reality promotes virtual environments with objects from the physical world shown in real time. For that, it's necessary to be concerned with the immersion of the user in this simulated environment, increasingly seeking to bring it closer to a possible desired reality. This paper proposes the development of image processing in order to perform the segmentation of images to identify what is foreground and background in order to facilitate the union of virtual and real images. Thus, the present work obtain real images of the user using the off-highway truck simulator CAT793F, through a camera, to be able to perform the segmentation of such images with artificial intelligence techniques.The convolutional neural network architectures "U-net" and "DeepLabV3+" are applied to perform image segmentation. As a result, metrics with around 90% accuracy were presented and and the best model was determined.

## 1. INTRODUCTION

Computer vision is an area of computing that has been continuously growing. Due to its immense usefulness in both everyday and industrial settings, this science is responsible for "machine vision" and employs artificial intelligence (AI) tools such as pattern recognition, image analysis, and semantic segmentation to generate its results.

As shown by Gonzalez-Sosa et al. (2020), most applications developed within computer vision were focused on direct or indirect interactions with cameras in third-person views (TPV). However, with the release of cameras and technologies like GoPro, there has been a significant increase in interest in the study of first-person views (FPV). Notable technologies addressing FPV in computer vision include Virtual Reality (VR). This technology creates environments with virtual scenes and objects, simulating a particular reality. These immersive technologies use hardware such as the "Oculus Rift" for user and environment immersion.

Regenbrecht et al. (2017) demonstrates that the environment generated by VR is entirely constructed by simulated objects, and mixed reality (MR) blends the physical and digital worlds by combining a virtual environment with real-world objects. Typically, simulated body parts such as arms, hands, and legs are introduced into simulated spaces to increase the user's level of immersion within the processed scene, but various objects can also be introduced, depending on the user's needs. Hardware like the "Oculus Quest" can generate such realities, differing from the Rift in that it has a camera that generates real-time images, which are processed to assist in building these environments.

The challenges faced in immersive environments (IE) revolve around user immersion, and attempts to keep the user as immersed as possible within this simulated reality (Gonzalez-Sosa et al., 2020).

The CAT793F is an off-highway truck manufactured by Caterpillar. Used in the mining sector, it is a high-priced asset that requires specific training,as it is large dimensions and is used in highly dangerous environments. The use of simulators allows for initial training and retraining of drivers in a controlled environment, where it is possible to simulate risk situations, assess and monitor failures and experience while driving.

This study aims to segment objects from the real world, such as the user's arms and the CAT793F off-highway truck simulator shown in Figure 1, to insert them into a virtual environment in real time, aiming to generate increasingly immersive MR that is closer to the reality of the physical world. For this work, two neural network models will be used to segment input images captured by a first-person view camera of users manipulating the off-highway truck simulator, in order to find the best result among those generated by the two networks.

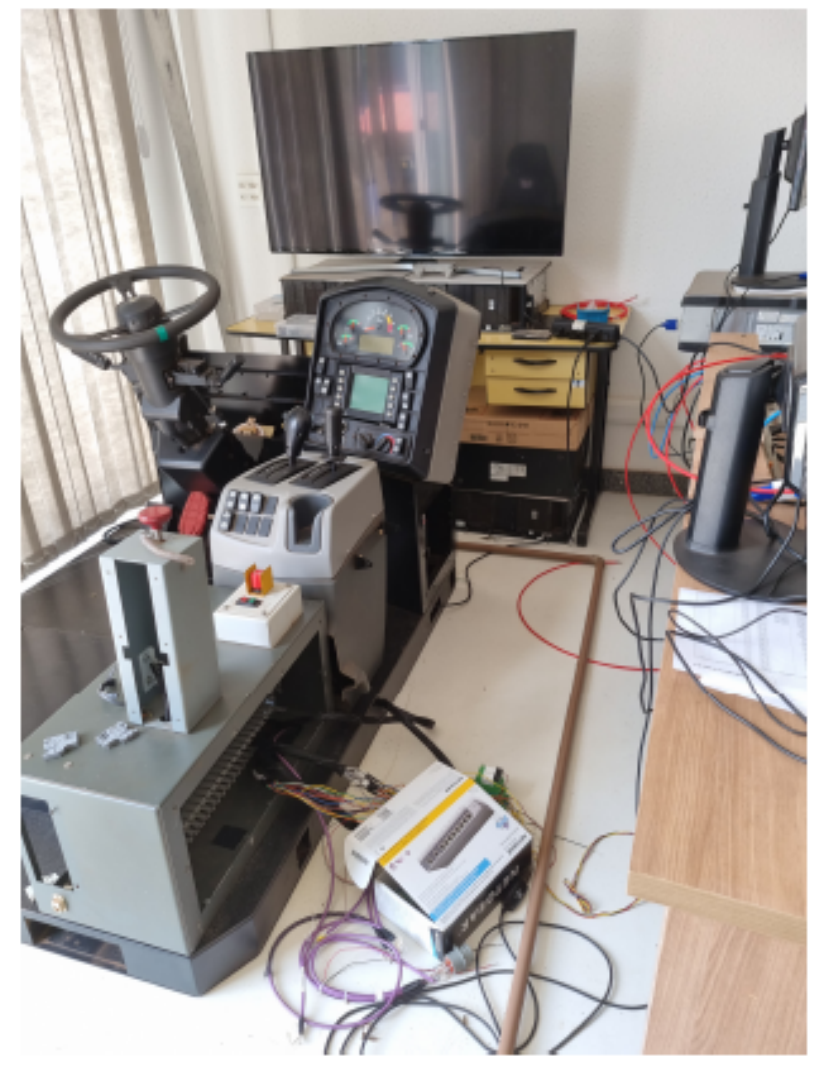

Figure 1. CAT793F off-road simulator

The rest of this paper follows this structure: In Section 2, the theoretical framework is presented, where some studies already conducted in the area are addressed. Next, in Section 3, the development is shown, highlighting its stages and showing the entire process to achieve the final goal. Then, in Section 4, the experiments conducted and their results are presented, and an analysis is made based on these results. Finally, in Section 5, the conclusion of the entire work is made.

## 2. RELATED WORKS

Distinct studies can be referenced around this topic. In this section, some studies are presented with the aim of providing a foundation for this work.

Zhang et al. (2021) begins their study by showing applications for image segmentation, such as in autonomous vehicles, medical diagnostics, and other applications. They demonstrate how CNN models differ for each case using metrics, and how this choice depends on experimentation. The author concludes the study by comparing metrics for a particular model in different scenarios, showing how each scenario requires a specific metric and model.

According to Zhang et al. (2019), with the advancement in the development of Convolutional Neural Networks (CNN), various Deep Learning models focusing on segmentation have been developed, some of which can be applied to problems of segmenting images with human body parts. The authors note that many studies focus on the accuracy of segmentations, and while some studies achieve rapid results, there is no balance between accuracy and speed. Results with high accuracies can be achieved, but GPU usage is required due to the high number of layers in neural networks, which requires a high level of processing and memory storage. However, real-time segmentation remains challenging. The study develops portrait image segmentation, focusing on human face images, in real-time using FCN (Fully Convolutional Network) models (Long et al., 2015), namely “PortraitFCN” and “PortraitFCN+”, both based on the “FCN-8s” model. The author uses these networks for segmentation and presents excellent results with four different datasets.

According to Pigny and Dominjon (2019), the insertion of user body parts, such as arms and legs, into simulated environments increases the user’s sense of presence and helps with spatial perception within the immersive reality, such as distance, depth perception, and positioning. The authors mention that many studies reconstruct the body through simulations within the virtual environment (VE) as one approach. However, these systems do not show many advantages in terms of simulation quality and may encounter problems in the process, hindering faithful results. Another approach is to mount a camera on the user’s head, close to their real-world vision (FPV), which records the objects to be introduced into the VE. This method is known as VST (Video See-Through) and is more efficient because, as it uses real images, it does not undergo object reconstruction steps for generation in virtual environments like the previous method, but rather through segmentation processes of these objects in the image using CNN.

The COCO dataset, which is a dataset available for semantic segmentation, was applied to train the chosen CNN model, which in this case is a modification of the “U-net”, in (Ronneberger et al., 2015). The authors refer to their developed model as “See-Through U-net” (SSTU-net). Images are collected using a ZED mini camera attached to the HTC VIVE headset. Finally, they use the trained network with the COCO dataset to segment the images generated by the input camera and show excellent segmentation results, both in the quality of generated images and in metrics and execution time.

Pigny and Dominjon (2019) also show that there are some conditions for reproducing this process, such as identifying human bodies as the same class, even if they appear very different from each other. The speed of the segmentation process is another condition mentioned, as it needs to keep up with the necessary update rate to maintain the immersion of the environment.

Gonzalez-Sosa et al. (2020) shows that one of the main problems in immersive environments is the “Sense of presence”, which revolves around the subjectivity of being in a remote environment without having moved in the physical world, a fact that interferes with the user’s perception of reality, thereby impairing the final outcome projected for mixed reality. The author proposes the segmentation of human arms for better immersion of the user in mixed reality and to solve the problem, uses a Samsung VR headset coupled with the Samsung-S8 cell phone, in front of a Chroma Key cloth, using an application to record videos with the phone’s camera in different situations and with different users. Two neural network models are proposed to perform segmentation of the acquired images, among them is “DeepLabV3+”. Finally, all the good results generated by the networks used by the author are shown.

Based on the studies presented, the present paper aims to study, implement, and analyze two models present in these studies, namely “U-net” and “DeepLabV3+”, in order to determine which of them best suits the segmentation problem of the user and the cockpit in off-highway truck simulation.

## 3. PROPOSED DEVELOPMENT

With the study and understanding of the problem, the development of the paper is proposed, following the flow shown in Figure 2, divided into 6 stages, which are detailed as following.

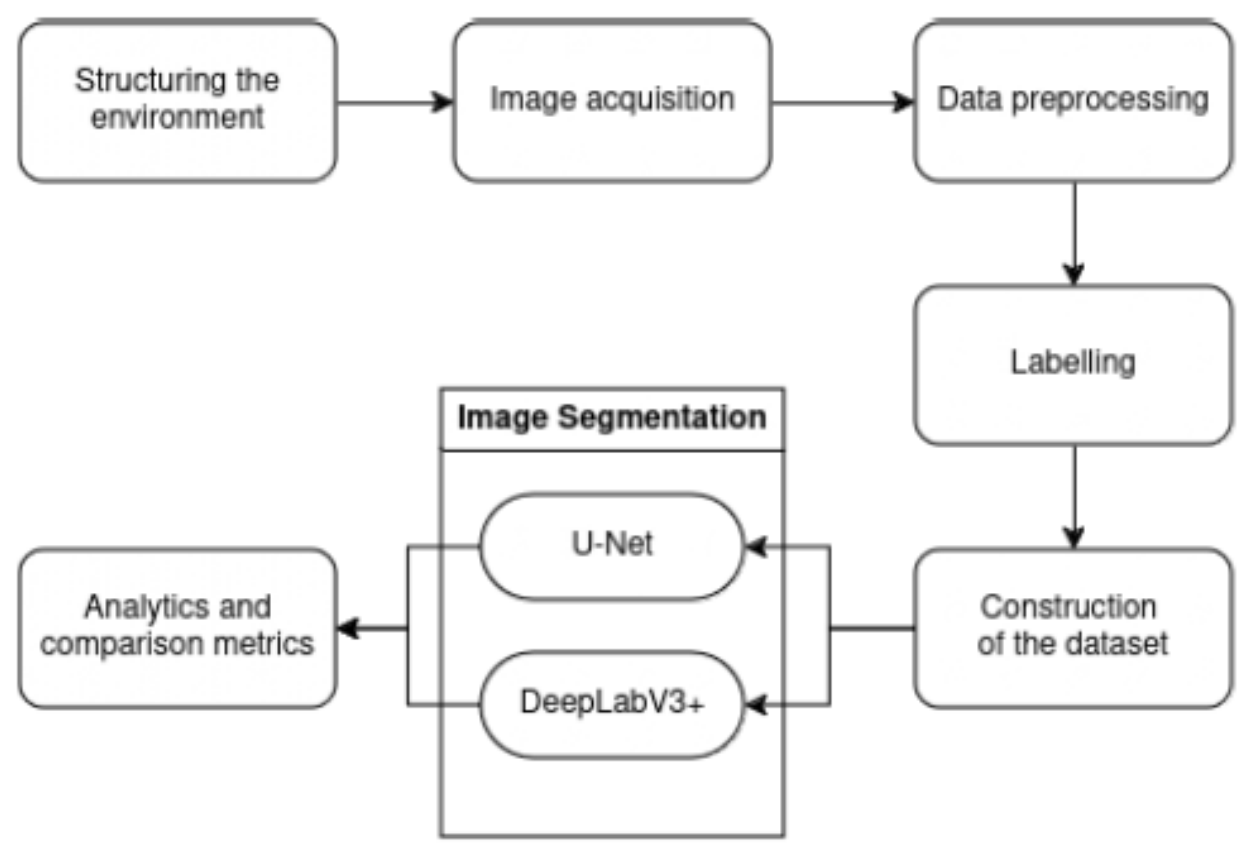

Figure 2. Development flowchart.

### 3.1 Structuring the environment

To carry out the work, the initial step was to acquire images in a controlled environment. For this purpose, it was necessary to use Chroma Key cloths to assist in the process of generating foreground and background images for neural network training. Subsequently, the construction of a structure that could support the fabric and envelop the entire CAT793F simulator was necessary to ensure correct image acquisition and facilitate image processing.

For the construction of the structure covering the CAT793F simulator, 7 PVC pipes of 6 meters in length and 25 mm in thickness, 4 cross fittings, 12 elbows, and 16 Ts were used. During the assembly stage, the pipes were cut and fixed in their respective positions. Subsequently, the covering stage with the Chroma Key cloth was initiated, starting with the sides, with the cloth being secured to the top using cloth clamps. Once the sides were completed, the top was covered using the same clamps that secured the sides. This led to the image acquisition stage.

### 3.2 Image acquisition

For image collection, a ZED 1 camera attached to a cap was used to simulate first-person view, closely resembling the simulated user's perspective. A total of 5 videos were recorded, featuring 4 different users. In these videos, users simulated interaction with the cockpit, moving their arms and heads to various positions, such as upwards and towards corners with high light incidence. This was done to facilitate testing during the segmentation stage. On average, the videos ranged from 40 to 60 seconds in duration. Figure 3 shows examples of images generated during the acquisition process.

### 3.3 Data preprocessing

In the data preprocessing stage, the aim is to enhance the images according to a specific pattern or objective, as images may have issues that hinder the segmentation process, such as corners with excessive light or oversaturation of color. To address these problems, it's necessary to apply filters that resolve the noise present in the input.

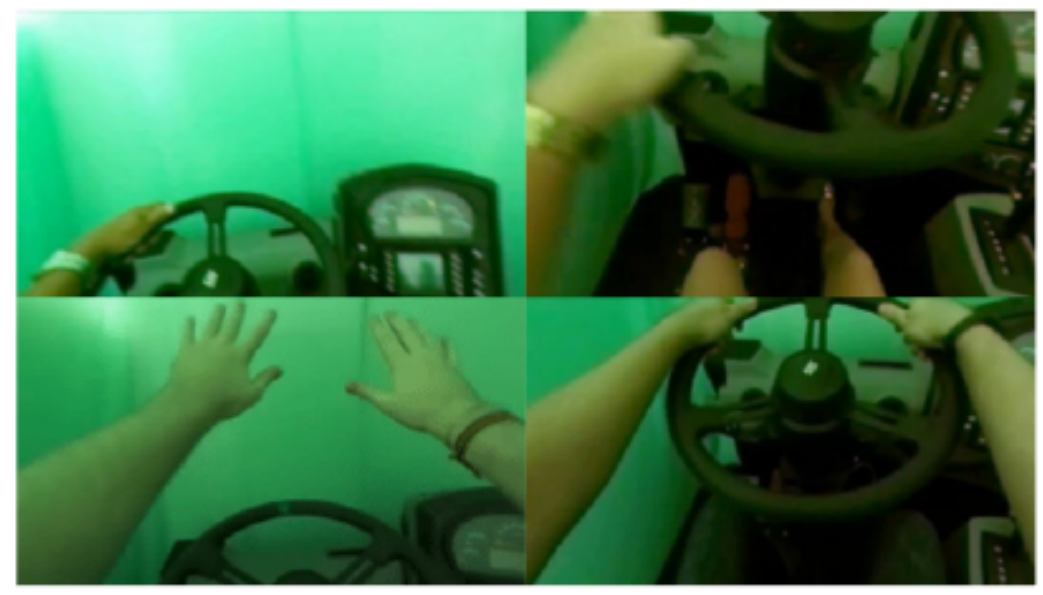

Figure 3. Frames taken from the image acquisition process.

With the collected images, color, saturation, and contrast corrections were applied to achieve a uniform background color and enhance the visibility of the arms and the simulator. This was necessary because the left corner of the cabin had more light incidence than the rest, and the green reflection was very noticeable on the arms, as seen in Figure 3. These corrections were made using Blender software, which is specifically designed for video and image editing and production. The "Color Correction" node was used in Blender to perform the corrections on the input videos. In this node, the key parameters for editing were saturation and contrast. The saturation parameter was reduced to achieve a satisfactory result, attempting to lighten the green and remove the green shadows present on the user's arms. Contrast values were increased to highlight the simulator and the arms against the green background. Finally, using the same software, the videos were separated into frames and added to the dataset.

### 3.4 Labeling

After performing the processing and separation of all collected images, the next step is labeling, where random images are selected to create binary masks, in black and white, which are used in the training and testing of the networks. For creating these masks, the Pixel Annotation Tool (Kläser, 2010) was used. In Figure 4, an example of the masks used in the work can be seen, constructed from the reference image.

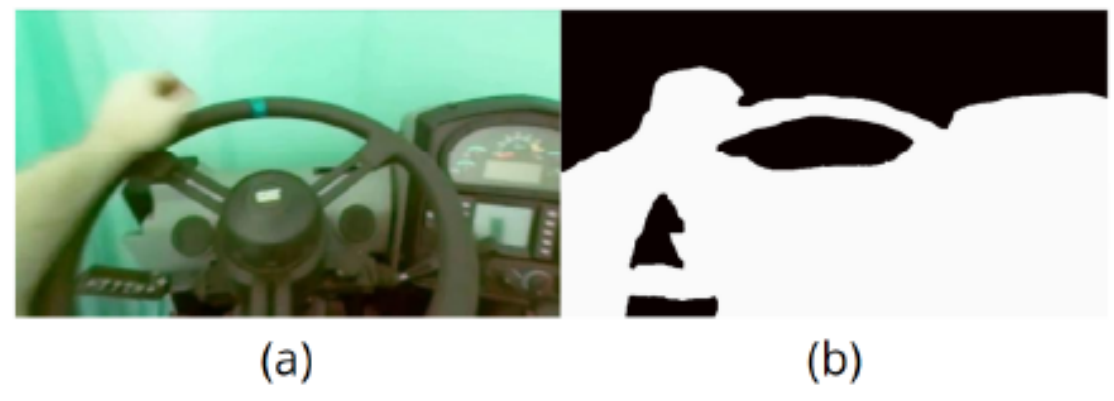

Figure 4. Input image (a), and corresponding mask (b).

### 3.5 Construction of the dataset

The dataset consists of a set of preprocessed images used in the image segmentation stages, more specifically provided to the neural network for processing. The videos, after undergoing image treatment, are divided into frames and separated for the selection stage. As shown in the labeling section, random images are selected to be used, and masks are also produced.

Then, the process for creating the database begins, where a Python script is executed for this purpose. This script standardizes the resolution of all input images and masks to 512 by 512 pixels and converts all images to the same format, which in this case is PNG. Finally, the code separates the input images and masks into their respective folders, “input” and “mask”. The final dataset consists of 232 input images and 232 masks.

### *3.6 Models applied for image segmentation*

In this stage, the semantic segmentation of the corrected images occurs through models of convolutional neural networks. Two network models were chosen for segmenting the images, namely the ”U-net“ (Ronneberger et al., 2015) and the ”DeepLabV3+“ (Chen et al., 2018), to obtain different parameters and results, thus enabling comparison and evaluation of the best solution for the problem.

*U-net:* This model was created and developed by Ronneberger et al. (2015) for the segmentation of biomedical images. This model is named after its U-shaped architecture, which consists of two pathways: the contraction pathway (Encoder) and the expansion pathway (Decoder), as shown in Figure 5.

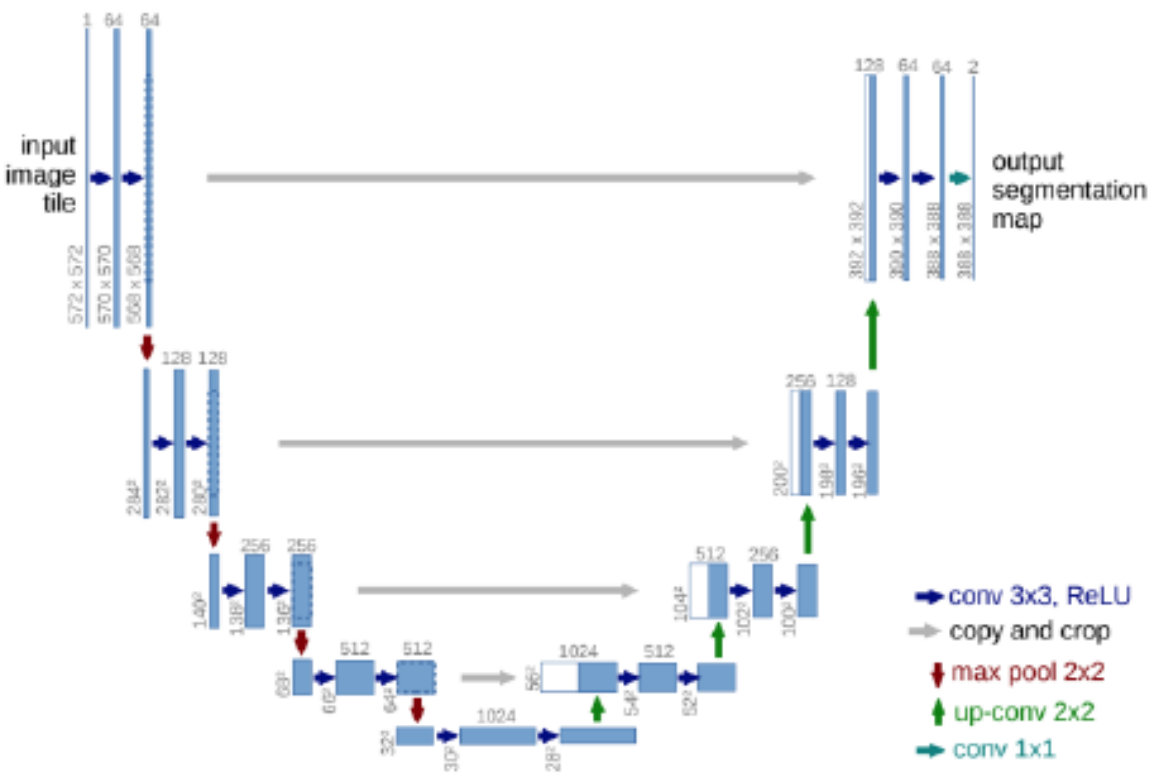

Figure 5. Architecture of the U-net model. Source: Ronneberger et al. (2015)

The contraction pathway follows a sequence of operations, starting with convolutions, an operation that correlates the input image with a specified kernel, multiplying pixel by pixel and summing them to generate the final image. In this case, 3x3 pixel kernels are used.

These convolutions are followed by activation functions, which are functions that decide whether or not the result of the layer will be used in the network. The function used in this case is ReLu (*rectified linear unit*), which returns 0 for negative values, and the value itself for positive input values, that is, it only activates positive elements. The ReLu function is represented by $ReLu(x) = max\{0, x\}$.

Following that, 2x2 Max pooling operations are performed, which basically involve reducing the resolution of the image by selecting the maximum value in the neighborhood of the specified regions, which in this case are 2 by 2 pixels.

In the expansion path, the image is increased to its original size. To achieve this, similar operations to the first path are used, employing 3x3 convolutions and activation functions (ReLu). Instead of the max pooling operation, transpose convolutions are used to increase the input resolution, thus proceeding to the final stage.

For the final stage of this process, a 1x1 convolution operation is used, followed by a sigmoid activation function, $sigmoid(x) = \frac{1}{1+exp(-x)}$. This function maps and transforms real values into outputs in the range 0 to 1.

Finally, it is possible to check the output as a binary mask of the input inserted into the network, and then performs the training and classification processes.

*DeepLabV3+:* The second chosen model was developed by Chen et al. (2018), which builds upon previous versions and enhances the last model called ”DeepLabV3”, aiming to construct an improved, faster, and more efficient network. This model utilizes the Encoder-Decoder architecture, as shown in Figure 6. Similar to the other model, it takes images and masks of the same size as input for training and testing.

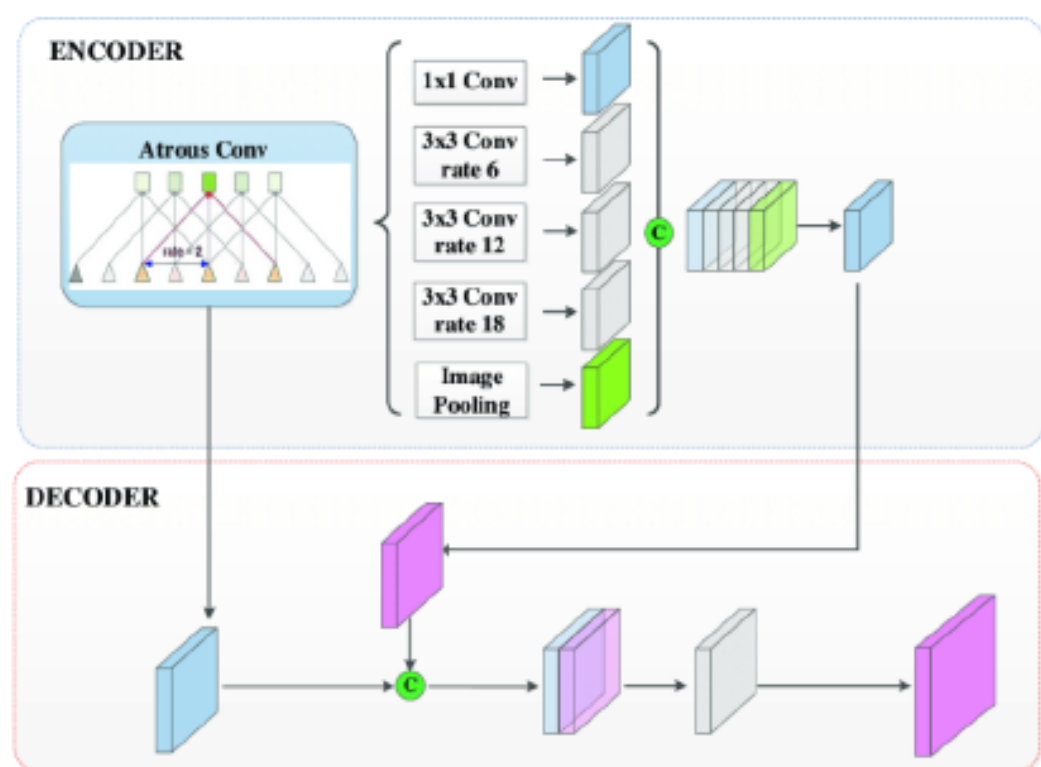

Figure 6. Architecture of the DeepLabV3+ model. Source: Chen et al. (2018).

The Encoder pathway in DeepLabV3+, utilizes the Atrous Spatial Pyramid Pooling (ASPP), which is an advancement of DCNN as its dilated convolution operations can have their rates determined separately, as seen in Figure 6. Then, the ASPP performs the pooling operation. Finally, a 1x1 convolution is performed to reduce the output dimension of the ASPP, forming the set inserted into the Decoder pathway of the structure. All convolution operations are followed by the ReLu activation function.

The decoding stage begins with a 1x1 convolution of the initial information to reduce the number of input channels, and this is concatenated with the encoding stage, which underwent a 4x upsampling, increasing its dimension. Next, two more operations are performed: a 3x3 convolution and another 4x upsampling, completing the structure. As with the previous model, the output is a binary mask of the input fed into the network.

*Training and testing approach:* A neural network, as shown in Shihab (2006), typically utilizes forward propagation and backpropagation as key tools for its learning process. In the propagation stage, the data passes through the layers of the network, in their respective nodes, with randomly predetermined weights. At the end of this process, the network can calculate the overall loss, indicating through this error value whether the network has performed well or not. Once this stage is completed, backpropagation begins. It selects the desired final values and compares them with the current outputs. With the help of an optimization function, it propagates this difference through the network, adjusting its weights accordingly.

The optimization functions play a fundamental role in this stage of training, as seen in Ribeiro et al. (2020). These functions have a significant influence on the performance of a neural network. They are responsible for reducing errors encountered by the networks, relative to the initially desired ones. In this paper, the Adam function, proposed by Kingma and Ba (2015), was used, which aims to minimize the error of each training epoch, that is, to improve the performance of the network.

During the training stage, hyperparameters are also defined, starting with the batch size, which divides the input dataset into batches, forcing the network to train with different sets of images at each iteration. Another defined parameter is the learning rate, which is a value ranging from 0 to 1 and controls how quickly the model adapts to the solution of the problem, as it is responsible for the step of updating weights in the nodes of the network. Finally, the number of epochs is defined, which consists of the number of times the network reproduces the entire training process. These vary for each network model due to their specificity for each case.

Both networks followed the same training pattern, using the preprocessed inputs and labeled masks inserted into the database, and going through the processes explained earlier. Following this, they moved on to the testing stage.

The testing stage verifies the effectiveness of the training performed, testing the network with a specific set of images, putting to the test the learning acquired by it in the previous stage. In this phase, evaluation metrics are defined to conduct an analysis with clarity and efficiency.

### *3.7 Analytics and comparison metrics*

For the purpose of analyzing and comparing all the obtained results and the performance of the network, metrics are used throughout the training and testing process. This enables us to understand the real functioning of the presented models in light of all the data and results presented.

Choosing the correct metric is crucial for conducting an effective analysis of the results obtained from the specified models. Hicks et al. (2022) discusses several metrics and based on that, the following metrics were chosen: loss function generated during the training phase, precision, recall, accuracy and IoU.

Accuracy (ACC) indicates an overall performance of the model, averaging correctly classified samples over the total number of samples. This metric is represented in Equation 1, where TP represents true positives, FP false positives, TN true negatives, and FN false negatives.

$$accurracy = \frac{VP + FP}{VP + FP + VN + FN} \tag{1}$$

The metric Recall (REC), or true positive rate, which can be seen in Equation 2, represents the network's ability to determine whether a true positive sample is correctly identified or not.

$$recall = \frac{VP}{VP + FN} \tag{2}$$

*Precision* is the metric that effectively shows the percentage of samples that were correctly identified. This metric can be seen in Equation 3.

$$Precision = \frac{VP}{VP + FP} \tag{3}$$

Finally, Zhang et al. (2021) highlights the importance of the *Intersection-Over-Union* (IoU) metric, also known as the *Jaccard* metric. This metric is known to be extremely efficient in evaluating binary image segmentation problems, which is the problem addressed in this work. IoU is calculated through the overlap area between the mask predicted by the network and the reference mask, divided by the union area of both masks, represented by the equation $IoU = \frac{Interserction}{Union}$.

## 4. EXPERIMENTAL RESULTS

This section presents the results and analysis of the results obtained in this work.

### *4.1 First experiments*

In the first experiments, we have the construction of the cabin to obtain the images, the collection of images, their pre-processing and the creation of the masks in the labeling stage, aims to generate the *dataset*.

For the construction stage of the cabin supporting the Chroma Key, the desired objective was achieved, considering that such a structure supported the weight of the cloth, both on the sides and on the top, covering all parts surrounding the simulator. As shown in Figure 7 (b), the supporting beams on the sides did not quite match the design due to a mistake in the pipe cuts, but this setback was overcome, and the final structure was satisfactory. The assembled final structure stood approximately 2 meters tall, 1.5 meters long, and 1.2 meters wide. Another satisfactory aspect achieved was the lighting; as the structure was set up near a window, the interior of the cabin was well-lit with daylight. The external part of the cabin, showing the previously described structure, can be seen in Figure 7 (b). Additionally, Figure 7 (c) displays the interior of the structure with the simulator.

In the data acquisition stage, as described earlier, 5 videos were recorded, with 4 different individuals. A frame from one of these videos can be seen in Figure 3.

Finally, the input images were processed and labeled. In Figure 8, a comparison between the untreated input (a) and the treated input (b) after color correction can be seen. An example of the labeling done is shown in Figure 4.

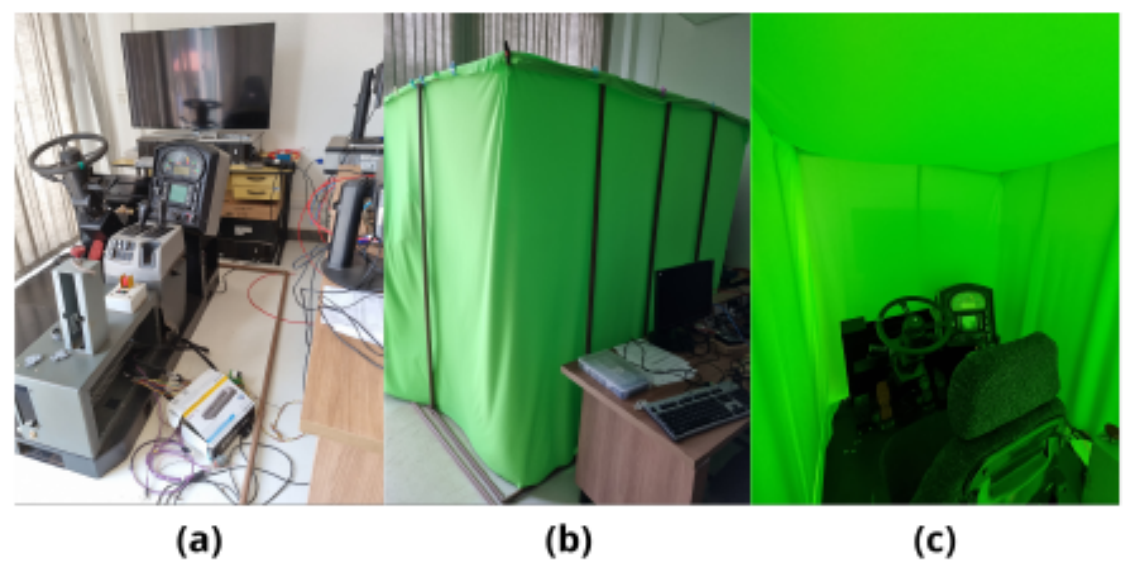

Figure 7. (a) CAT793F simulator, and the external (b) and internal (c) parts of the constructed structure.

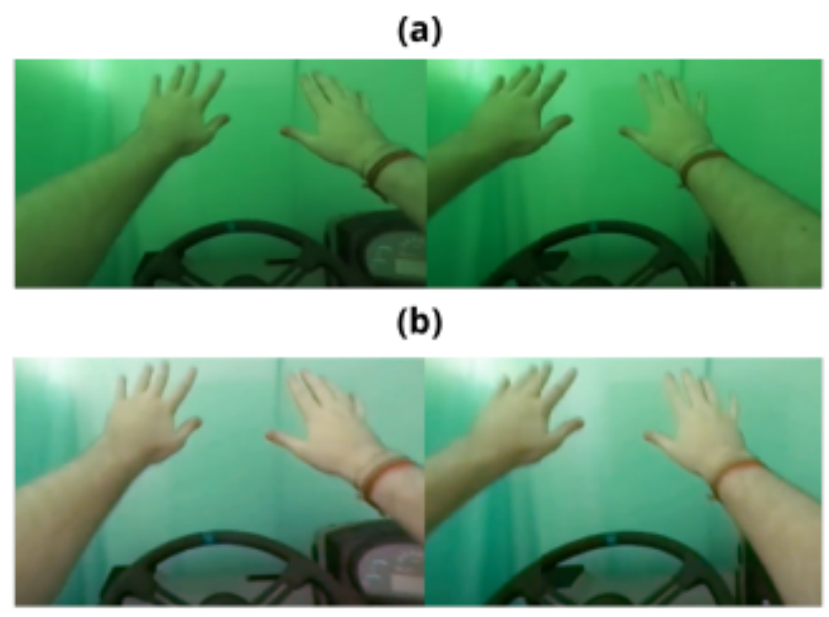

Figure 8. Original untreated image (a) and image treated with color correction (b).

### *4.2 Hyperparameters configuration*

In this section, the hyperparameters used for the construction, execution, and finalization of the specific CNN models are presented. The architectures were used in their original structures; only the hyperparameters were modified and metrics were chosen for analysis. The hyperparameters chosen for both models were **Learning Rate** $= 1e{-}4$, **Batch size** $= 2$ and **Epochs** $= 50$.

In addition to the hyperparameters, it is noteworthy that the Adam function was used as the optimization function for both models. Finally, 4 metrics were chosen for model evaluation, namely those explained earlier in Section 3.7: ACC, recall, precision, and IoU.

### *4.3 Results obtained through the proposed models*

Here, the results obtained from the training and testing of both CNN models proposed in this work are presented. For training the networks, 75% of the dataset, totaling 232 images, were randomly selected, resulting in 174 images. The remaining 25%, 58 images, were allocated for the validation stage. This separation is done to avoid overfitting, which is a biased and skewed result.

In Figure 9, some of the results generated by the two CNN models developed in this work are shown. The input image, its original mask, and the output of both networks are displayed, respectively. It is possible to visually assess the difference in the quality of the results between the two networks. These differences are discussed in Section 4.4.

### *4.4 Analysis of results*

The first parameter to be analyzed is the model loss function curve (Figure 10). As shown previously, the loss function generates an error at each training epoch, and this function seeks to reduce this error at each iteration. From the analysis of the two curves, it is possible to see how the U-NET model learns faster. The smoothing of this curve in fewer periods demonstrates this fact.

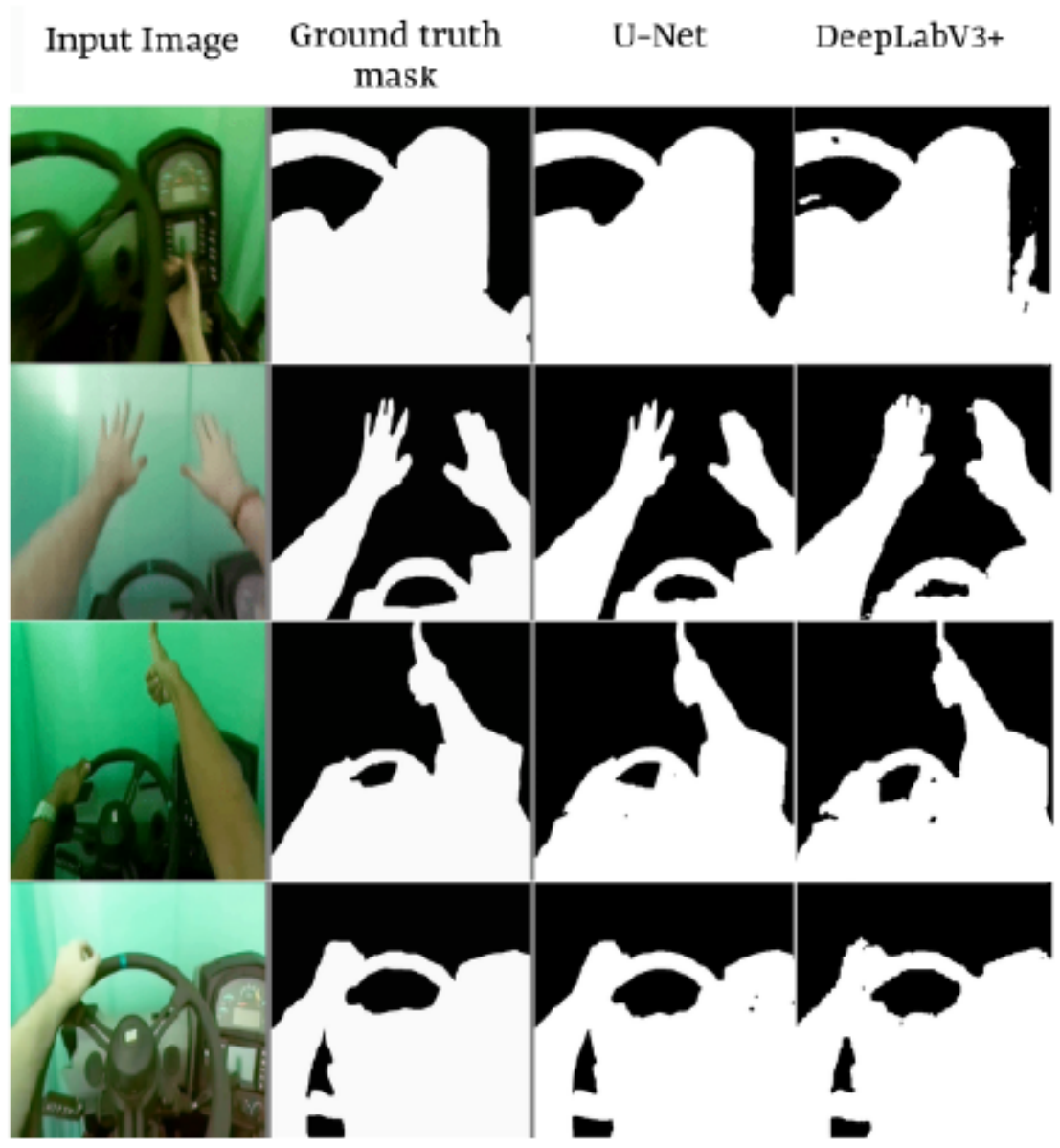

Figure 9. Qualitative comparison of results.

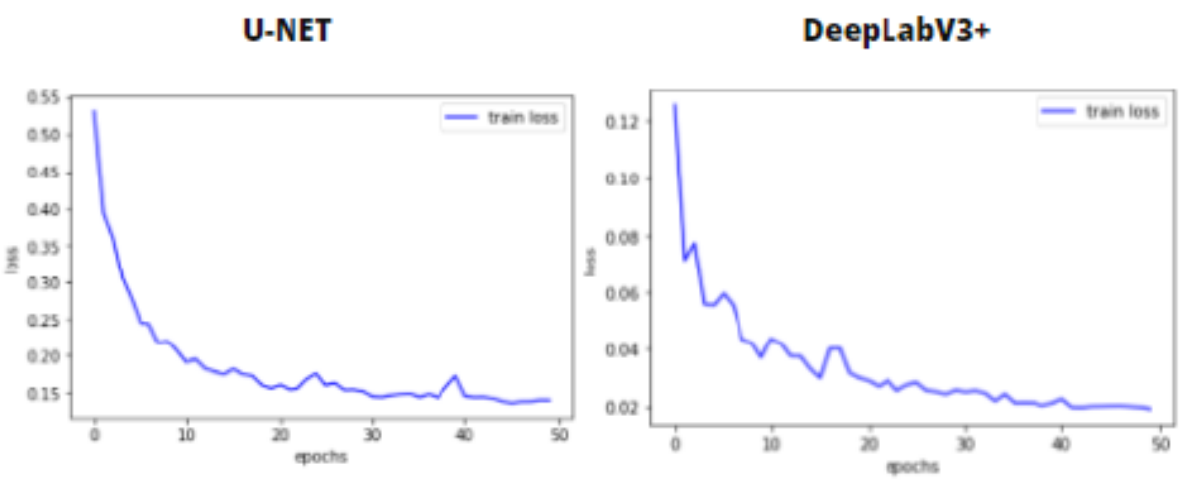

Figure 10. Loss function curves of the studied models.

It is also important to analyze the models in relation to execution time. In Table 1, it is possible to check the execution time values in seconds for each model, being the average time per epoch, the total training time and the total testing time. The U-NET model stands out in all time parameters, having more efficiency in processing time compared to the second model.

Table 1. Processing time for the models.

| Time (s) | Unet | DeepLabV3+ |
|---:|---|---:|
| Average time per epoch | 23.2 | 74 |
| Average time for learning | 1780.6 | 3762.4 |
| Total time to test | 13.32 | 24.26 |

Finally, the four highlighted metrics are analyzed. Table 2 shows the comparison between the two models regarding the averages of the ACC, *recall*, *precision* and IoU metrics. It is notable how the first model stands out in all parameters, having better values compared to the second model. The IoU metric is highlighted, which presents approximately 5% difference between the two networks.

Table 2. Quantitative comparison of models.

| Metric | Unet | DeepLabV3+ |
|---|---|---|
| Accuracy | **98.69%** | 96.34% |
| Recall | **98.39%** | 96.76% |
| Precision | **97.57%** | 95.92% |
| IoU | **96.05%** | 91.02% |

### *4.5 Application of results*

Finally, based on the analysis performed and according to all the observed parameters, it was concluded that the best CNN model was the U-NET architecture. A final study was then carried out, using this model to train and test the environment without the presence of *Chroma Key*, thus making learning the network difficult, due to the vast number of different elements present in the image. The *dataset* used has 80 images in total, 60 of which were used for the learning stage and 20 for the testing stage.

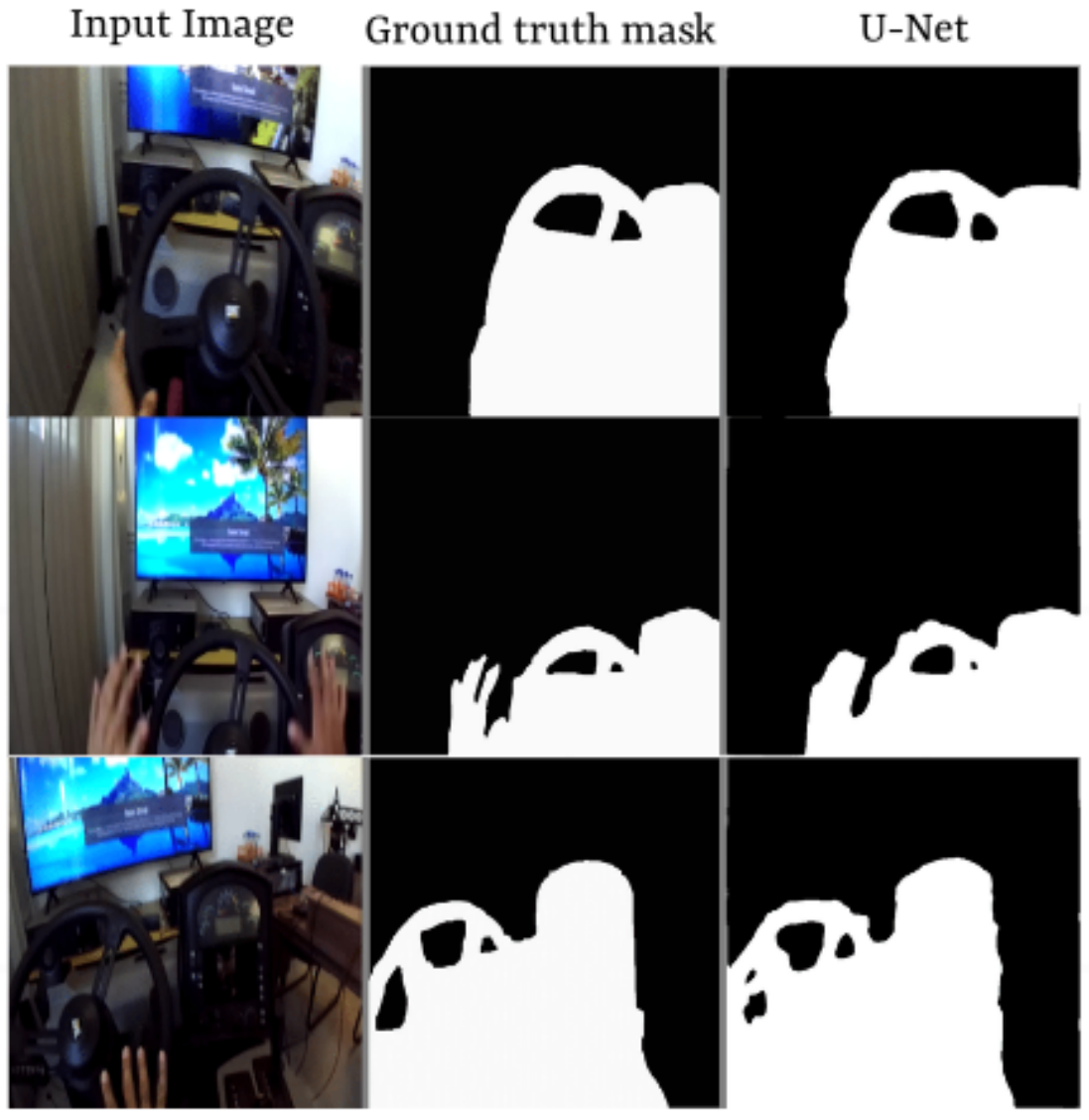

Figure 11. Analysis result without the aid of *Chroma Key*

Figure 11 shows a set of images that show the results obtained by U-NET for this case. The efficiency of the CNN is visually noticeable, which even without the help of *Chroma Key* can obtain results close to desired ones.

Table 3. Metrics without *Chroma Key*

| Metrics | Unet |
|---|---|
| *Accurracy* | 91.56% |
| *Recall* | 94.74% |
| *Precision* | 92.12% |
| IoU | 87.67% |

In Table 3 it is possible to check the metrics generated in the validation stage of the study carried out without the aid of *Chroma Key*. It can be seen that these do not differ significantly from those produced by the study with the help of *Chroma Key* as shown in Table 2. Finally, it is possible to highlight the IoU metric, with approximately 87% assertiveness.

## 5. CONCLUSION

From the results presented, it is possible to conclude that the choice of neural network architecture obviously directly influences the results of each situation studied. In this case, as shown and analyzed, the U-NET architecture proved to be more efficient both in results and metrics and in processing time, for the case studied.

To continue the study, this model will be applied to the complete videos, allowing for more complex and detailed analyses, thus approaching the problem from another parameter. The aim is also to prepare the network for the virtualized environment, which will be able to segment images in real time and insert them into the given reality.